\documentclass{article}

\usepackage{arxiv}

\usepackage[utf8]{inputenc} 
\usepackage[T1]{fontenc}    
\usepackage{hyperref}       
\usepackage{url}            
\usepackage{booktabs}       
\usepackage{amsfonts}       
\usepackage{nicefrac}       
\usepackage{microtype}      
\usepackage{lipsum}
\usepackage{graphicx}
\graphicspath{ {./images/} }

\usepackage{multirow}
\usepackage{subfigure}
\usepackage{hhline}
\usepackage{amsmath}

\title{Predicting Landfall's Location and Time of a Tropical Cyclone Using Reanalysis Data}

\author{
 Sandeep Kumar \\
   Department of Mathematics, Shaheed Bhagat Singh College,\\ University of Delhi.\\ \& \\
  Department of Computer Science, IIIT Delhi\\ 
  New Delhi, India. \\
  \texttt{ sandeep\_kumar@sbs.du.ac.in, sandeepk@iiitd.ac.in} 
   \And
    Koushik Biswas\\
  Department of Computer Science \\
  IIIT Delhi\\
  New Delhi, India, 110020 \\
  \texttt{koushikb@iiitd.ac.in} \\
   \And
 Ashish Kumar Pandey \\
  Department of Mathematics\\
  IIIT Delhi\\
  New Delhi, India, 110020 \\
  \texttt{ashish.pandey@iiitd.ac.in} \\
}

\date{}

\begin{document}
\maketitle
\begin{abstract}
Landfall of a tropical cyclone is the event when it moves over the land after crossing the coast of the ocean. It is important to know the characteristics of the landfall in terms of location and time, well advance in time to take preventive measures timely. In this article, we develop a deep learning model based on the combination of a Convolutional Neural network and a Long Short-Term memory network to predict the landfall's location and time of a tropical cyclone in six ocean basins of the world with high accuracy. We have used high-resolution spacial reanalysis data, ERA5, maintained by  European Center for Medium-Range Weather Forecasting (ECMWF). The model takes any 9 hours, 15 hours, or 21 hours of data, during the progress of a tropical cyclone and predicts its landfall's location in terms of latitude and longitude and time in hours.  For 21 hours of data, we achieve mean absolute error for landfall's location prediction in the range of 66.18 - 158.92 kilometers and for landfall's time prediction in the range of 4.71 - 8.20 hours across all six ocean basins. The model can be trained in just 30 to 45 minutes (based on ocean basin) and can predict the landfall's location and time in a few seconds, which makes it suitable for real time prediction.
\end{abstract}


\section{Introduction} \label{intro}
Predicting natural disasters is one of the difficult prediction problems because of the complex interplay between various cause factors, which vary with space and time. One such natural disaster is a Tropical Cyclone (TC) that frequently occurs in tropical and subtropical regions of the world. TCs are also called Hurricanes or Typhoons in different parts of the world. TCs are characterized by low-pressure areas with an atmospheric circulation that brings heavy rainfall, strong winds, thunderstorms, and flash floods in the coastal regions, thereby affecting human lives, property, transportation, businesses, and society. Each year TCs are responsible for the deaths of hundreds of people and billions of economic losses \cite{Grinsted23942, wcms2010teiofi}.  For example, recent cyclones Willa (2018) in the East Pacific (EP) ocean, Yutu (2018) in the East Pacific (EP) ocean, Harold (2019) in the South Pacific (SP) ocean, Amphan (2020) in the North Indian ocean, Marcus (2018) in the South Indian(SI) and Michael (2018)  in North Atlantic ocean caused economic losses of nearly US\$41 billion in total. 

The most important event during the progress of a TC is its landfall that is when it reaches land after crossing the coast of the ocean. The economic and human losses caused by a TC are centered around few kilometers of its landfall location. Therefore, it is crucial to predict the location and time of the landfall of a TC well advance in time with high accuracy. When a TC is over the ocean, the TC's real-time data may not be continuously available because of measurement constraints. Therefore, the landfall prediction model must be flexible regarding the data it requires for prediction. The model should pick continuous data from the course of the TC whenever available and provide predictions of landfall's location and time with high accuracy. In this study, we have used a deep learning model to predict the landfall's location and time of a TC. The model takes 9 hours (h), 15h, or 21h of continuous data, anytime during the course of a TC, and predicts the landfall's location and time at least 12h before the actual landfall time. The model performance is reported for six ocean basins - North Indian (NI), South Indian (SI), West Pacific (WP), East Pacific (EP), South Pacific (SP), and North Atlantic (NA) oceans. As per our knowledge, this is the first work that directly focuses on predicting the characteristics of the landfall of a TC using reanalysis data. Predicting landfall's characteristics is a challenging problem to deal with, as discussed in \cite{LEROUX201885}.

There are numerous existing TC track prediction models that can be classified into numerical (or dynamical) models, statistical models, and ensemble models \cite{nhcnoaa}. The numerical models rely on physical equations governing atmospheric circulations to capture the evolution of the atmospheric fields. These methods are computationally involved and need large supercomputers. The few major operational numerical models for track prediction are - European Center for Medium-Range Weather Forecasts (ECMWF), Global Forecast System (GFS),  Hurricane Weather and Research Forecasting model (HWRF), and Hurricane Multi-scale Ocean-coupled Non-hydrostatic model (HMON). The statistical models \cite{ha00310j} do not require high computational resources and rely on finding a relationship between historical data and cyclone specific features. The primary operational, statistical models for track prediction are -  Climatology and Persistence model (CLIPER5) and Trajectory-CLIPER. The ensemble models combine numerical and statistical models for prediction. Generally, ensemble models perform better than individual models \cite{tczt2000mefw}. Numerical methods and statistical methods have their own limitations, and we need to make a trade-off between computational cost and capturing of the complex relationship between various cause factors.

Recently, with the increase in the data related to tropical cyclones, various studies have appeared that have successfully applied machine learning-based models to predict various characteristics of a tropical cyclone \cite{LEROUX201885, mmgmha2016srnnt} which we discuss in detail in the next section. 
The article is organized as follows: Section~\ref{related} describes the related work in this direction, Section~\ref{data1} describes the data used in this study, Section~\ref{model} describes the deep learning model used to forecast the above mentioned prediction problem. Finally, in Section~\ref{results}, we present the results and analysis of our model. In Section~\ref{con}, we conclude and provide future directions. 

\section{Related Work}\label{related}

Initial studies regarding tropical cyclones track and intensity forecasts used Artificial Neural Networks (ANNs) \cite{csbdg2017swnnf, KOVORDANYI2009513}. Since the prediction problems related to atmospheric conditions involve both spatial and temporal components, deep learning models like Recurrent Neural Networks (RNNs), Long Short-Term Memory (LSTM) networks,  Convolutional Neural Networks (CNNs), and combinations of these have been successfully deployed to capture the complex non-linear interplay between various atmospheric components both of spatial and temporal nature. In \cite{mmgmha2016srnnt}, sparse RNN with flexible topology is used for the prediction of hurricane trajectory in the Atlantic ocean. In \cite{khjykkp2019dtfe}, ConvLSTM is used to predict the hurricane trajectory from the past density maps of hurricane trajectories. In \cite{abjags2018phturnn}, authors have presented a fully connected RNN model to predict cyclone trajectories from historical cyclone data in the Atlantic ocean. In \cite{gspymjzsz2018anmpttlstm}, a nowcasting model is presented based on an LSTM network to predict typhoon trajectory. 

Recently, few studies have dealt with TC formation, track, and intensity prediction problems using reanalysis data \cite{gsygk2020tctufl, boussioux2020hurricane, crxwxac2019ahcltff}. In \cite{crxwxac2019ahcltff}, reanalysis dataset has been used to forecast typhoon formation forecasting in NA, EP, and WP oceans. In \cite{gsygk2020tctufl}, authors have used historical data of a TC along with reanalysis data from ERA-Interim \cite{erainterim} to predict the track of TC with a lead time of 24h in six ocean basins. They propose a fusion network in which the output of CNN trained on wind fields and pressure fields from reanalysis data and output of an ANN trained on historical TC data are fed into another ANN network. Their model does not take the temporal aspect into account as they have stacked input from two-time steps $t$ and $t-6$ to feed into a CNN. In \cite{boussioux2020hurricane}, TC intensity and track prediction task is achieved with a lead time of 24h, using reanalysis data ERA5 \cite{era5}, historical TC data, and output from operational forecast models for NA and EP ocean basins. They have proposed framework Hurricast (HURR) consisting of seven different models that used different combinations of CNN, GRU, Transformers, and XGBoost models. They used data between $t$ to $t-21$ hours and capture the spatial and temporal aspects of data, thereby addressing the shortcoming of \cite{gsygk2020tctufl}. As there is no existing work that predicts the landfall characteristics from reanalysis and historical TC data, we will try to compare our model with the models presented in \cite{gsygk2020tctufl, boussioux2020hurricane}.


\section{Data}\label{data1}

\begin{figure*}[!h]
    \centering
    \includegraphics[width = 16.5cm, height = 6.5cm]{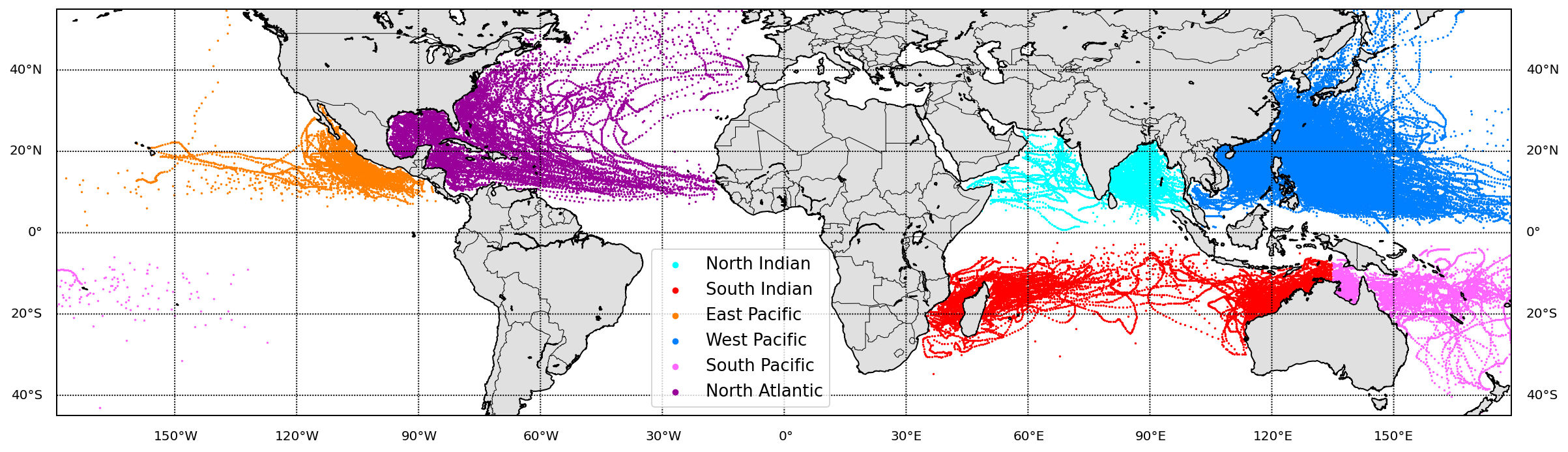}
    \caption{Trajectory of all cyclones till landfall in six ocean basins}
    \label{data}
\end{figure*}

\begin{figure}[t]
    \centering
    \includegraphics[width = 8cm, height = 8cm]{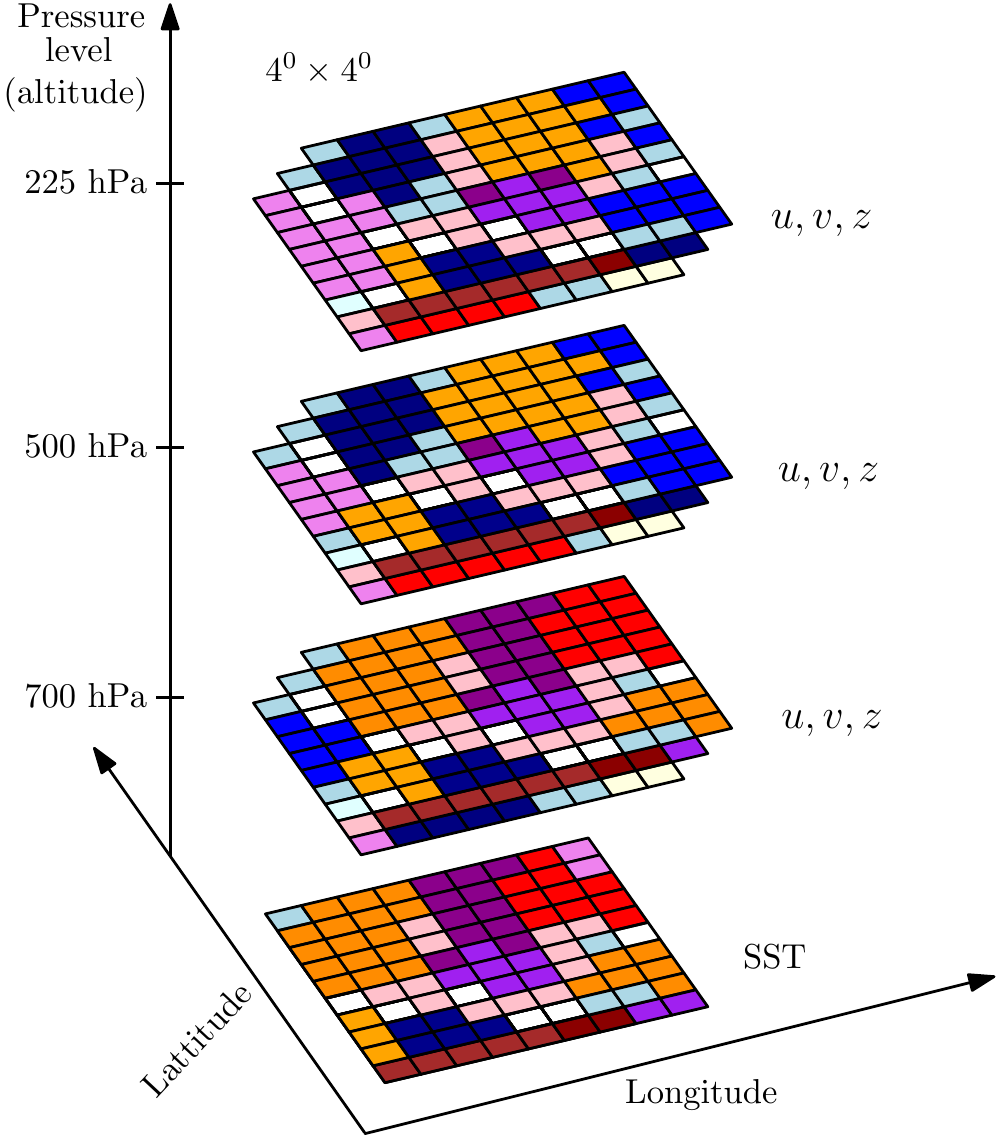}
    \caption{A pictorial depiction of u, v, z wind fields and SST}
    \label{datauvz}
\end{figure}

In this study, we have used two open-source datasets - historical cyclones data and reanalysis data. The historical track data is taken from NOAA database IBTrACS version 4 \cite{ibtrackdata, IBTrACS}. The dataset contains information about ocean basin, latitude, longitude, Estimated Central Pressure (ECP), Maximum Sustained surface Wind Speed(MSWS), distance to land, distance and direction of TC movement, etc at an interval of 3hours. From these features, we choose latitude, longitude, and distance to land for our study and exclude features like MSSW, ECP, distance, and direction of TC movement, which are used by two related works \cite{gsygk2020tctufl, boussioux2020hurricane}. As we are predicting the landfall's location and time of a TC, our dataset consists of all those cyclones which hit the coastal region, and for all such cyclones, only data corresponding to the time points when a TC was moving over the ocean is taken into account. If a cyclone moves from ocean to land and then from land to ocean and continues like this during its course, then the data corresponding to its presence over land is not considered. Also, if a cyclone moves from ocean to land and land to ocean multiple times, then each such movement is treated as a separate cyclone while preparing the dataset. Because of this process, our dataset size is shortened. We are not using features like  MSSW, ECP, distance, and direction of TC movement, which have lots of missing values to avoid further shortening of our dataset. We extracted data from 1981 to 2020 provided at an interval of three hours. The trajectory of all TCs till their landfall considered in this study are shown in Figure~\ref{data}.

The large scale atmospheric circulation of wind at different pressure levels plays a crucial role in determining the track of a TC. To capture this information, we have used ERA5 \cite{era5} reanalysis data produced by ECMWF in near to real-time. ERA5 is a fifth-generation reanalysis data covering global climate and weather since 1950. From 1979 onwards, the high-resolution data ERA5 replaces the ERA-interim. Reanalysis is a scientific way of producing globally complete and consistent data by gathering the information from various resources and validating them using the laws of physics. We extracted the $u, v$ components of wind and  $z$ geopotential  fields at three atmospheric pressure levels (225hPa, 500hPa, 700hPa) for a spatial extension of 4 $\times$ 4 degree and spatial resolution of 0.25 $\times$ 0.25 degrees (resulting in a grid of size 33 $\times$ 33), centered at the current TC location. The $u$ and $v$ component of wind represents its eastward-westward and northward-southward movement respectively. The $z$ geopotential represents gravitational potential energy of a unit mass relative to sea level. The choice of these variables is inspired by \cite{gsygk2020tctufl, boussioux2020hurricane} where authors have used values of these three variables at mentioned three pressure levels for a spatial extension of 25 $\times$ 25 degrees and spatial resolution of 1 $\times$ 1 degree (resulting in a grid of size 25 $\times$ 25). On the world map, one degree approximately equals 110 kilometers (KM). This way, we are utilizing the mentioned variables values for a spatial spread of around 440 KM (in comparison of 1320 KM by earlier two studies) with a spatial resolution of around 27.5 KM (in comparison of 110 KM by earlier two studies) around the TC center. Apart from these three variables, we extracted the sea surface temperature (SST) for the 33 $\times$ 33 grid centered at the TC location for each time point. A pictorial depiction of reanalysis data is shown in Figure~\ref{datauvz}.

\begin{table}[!h]
\centering
\caption{Dataset size and Landfall time (hours)}
\label{datasetDes}
\begin{tabular}{|p{1cm}|p{1.2cm}|p{1.2cm}|p{1.2cm}|p{1.2cm}|p{1.2cm}|}
\hline
\textbf{Ocean Basin} & \textbf{No. of TCs} & \textbf{Size of Dataset} & \textbf{Average Time} & \textbf{Min Time} & \textbf{Max Time} \\
\hline
NI & 205 & 5920 & 95.61 & 21 & 270 \\
\hline
SI & 282 & 10600 & 121.74 & 21 & 516 \\
\hline
EP& 116 & 4000 & 112.44 & 21 & 315 \\
\hline
SP & 189 & 5674 & 99.06 & 21 & 513  \\
\hline
WP & 1064 & 39166 & 119.43 & 21 & 606\\
\hline
NA & 401 & 11386 & 94.18 & 21 & 531\\
\hline
\end{tabular}
\end{table}

As we are using CNN, to feed the current location information, we have created two more 33 $\times$ 33 matrices $\operatorname{lats}$ and  $\operatorname{longs}$ for each time point of a TC. The each row of $\operatorname{lats}$ is equal to vector $(\operatorname{lat} +  0.25*k | -16 \leq k \leq 16 )$ and each column of $\operatorname{longs}$ is equal to vector $(\operatorname{long} +  0.25*k | -16 \leq k \leq 16 )$ where $(\operatorname{lat}, \operatorname{long})$ denotes the latitude and longitude of TC's current location. Feeding this information in CNN will enable it to generate distance and direction like features between two successive time points of a TC.

In Table~\ref{datasetDes}, the dataset size along with average landfall time after the initiation of a TC in six ocean basins are shown for all TCs with the minimum time difference between TC formation and its landfall as 21h.

\subsection{Training Dataset Preparation}

Let $T$ be the number of continuous data points (that is $3(T-1)$ hours of data) taken in the model to predict the target. For a fixed cyclone, let $T_L$ be the number of data points between cyclone formation and its landfall. For this TC, we created $T_L-(T-1)-3=T_L-T-2$ inputs, where a single input is a sequence of $T$ vectors of the form:

$$(\operatorname{lats}(t), \operatorname{longs}(t), \operatorname{u225}(t), \operatorname{v225}(t), \operatorname{z225}(t), \operatorname{u500} (t), \operatorname{v500} (t), \operatorname{z500}(t), \operatorname{u700} (t), \operatorname{v700}(t),\operatorname{z700} (t), \operatorname{SST}(t))$$

where $k\leq t\leq T+k-1$ and $k$ varies from $1$ to $T_L-T-2$. The target variables for each input are latitude and longitude at landfall or time (in hours) remaining to landfall of the cyclone from the current time $t$. One must note that by following the above process, we are predicting our target at least 12h before the landfall. For example, BELNA cyclone formed at 00 hours on 05 December 2019 in EP ocean, and the landfall happened at 15 hours on 09 December 2019, that is $T_L = 37$. Suppose $T= 8$, then this TC will generate $37 - 8 -2 = 27$ training data points. The collection of all such inputs across all TCs for a particular ocean basin will form the training dataset.


\section{Model Implementation and Training}\label{model}

As we are dealing with a dataset with both spatial and temporal dimensions, we have used a combination of CNN and LSTM models. In this section we will briefly describe the ANN, RNN, LSTM and CNN models. 

\subsection{Artificial Neural Network}

\begin{figure}[!h]
\centering
\includegraphics[width=0.6\textwidth]{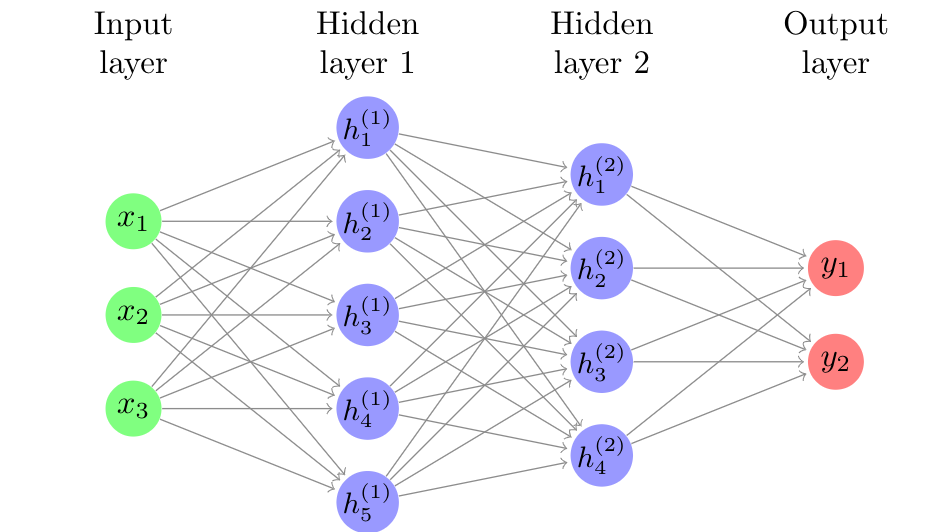} 
\caption{A fully connected ANN}
\label{annfig}
\end{figure}

ANNs \cite{mcculloch1943logical}  are connected networks of layers, where each layers consists of nodes. Generally the nodes in one layer are connected to nodes in succeeded and preceded layers. Each such connection is assigned weights, that regulates the information flow between nodes from one layer to another layer. The information flow at one node is a composition of a non-linear function with weighted linear sum of incoming connections. The non-linear function is called an activation function, which helps in learning non-linear relationship between input and output variables. The weights assigned to each connection is updated in a way to minimize the suitably chosen loss function through the Gradient Descent algorithm \cite{kiefer1952}. The intermediate layers between the input and output layers are called hidden layers. A fully connected ANN with two hidden layers is shown in Figure~\ref{annfig}.

Mathematically, the information flow in an ANN can be represented as $h_j^{(i)} = \sigma(W_j^{(i)} X)$, where $X$ is the input matrix containing bias, $h_j^{(i)}$ represents the information at the $j^{th}$ neuron in the $i^{th}$ layer, $W_j^{i}$  is weight matrix that determines interaction between nodes at $(i-1)^{th}$ and $i^{th}$ layer, and $ \sigma $ is the activation function. ANN models are not best choice for time-series data, where the output is a function of sequence of inputs. RNN models are best suited for time-series data that has ability of information persistence. 

\subsection{Recurrent Neural Network}

\begin{figure}[!h]
\centering
\includegraphics[width=0.6\textwidth]{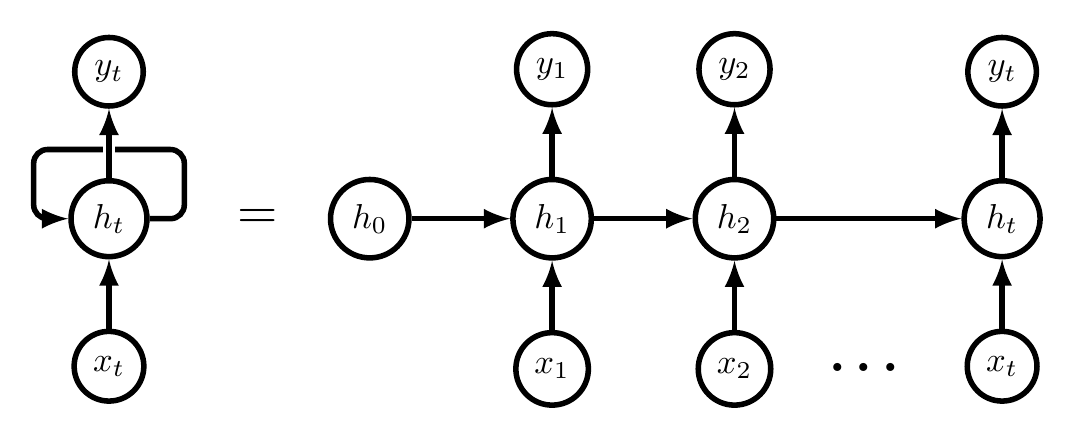} 
\caption{Structure of RNN}
\label{rnnfig}
\end{figure}

Recurrent neural networks (RNNs) \cite{rnn1, rnn3, rnn4}  are like ANNs with internal connections. They are suitable for time series data as they can remember information from inputs at past time points along with current input. A RNN structure is described in Figure~\ref{rnnfig}. Mathematically, the information flow in a RNN can be represented as
\[
h_t=\sigma(W_hx_t+U_hh_{t-1}+b_h), \quad y_t= \sigma(W_yh_t+b_y) 
\]
where $x_t$ is the input, $h_t$ is the hidden state, $y_t$ is the output at time $t$, $W_h, W_y$ and $U_h$ are weight matrices, $b_h, b_y$ are the biases, and $\sigma$ is the activation function.

However, they suffer from the problem of exploding and vanishing gradients \cite{279181}. LSTM networks \cite{lstm0,lstm1, lstm2, lstm3} overcomes this problem by maintaining three inner cell gates and a memory cell to handle past dependencies. LSTM are successfully applied on TCs track and intensity related prediction problems in \cite{abjags2018phturnn, gspymjzsz2018anmpttlstm, mmgmha2016srnnt}.

\subsection{Convolutional Neural Network (CNN)}

In recent times, CNNs \cite{6795724, kasihg2012icdcnn, 7785132} have become the default choice to deal with spatial data like satellite images or image-like data. They are inspired by the cortex visual system, where each neuron process data of its neighboring fields. Like an ANN, CNN also have an input layer, an output layer, and multiple hidden layers consisting of convolutional layers, pooling layers, and fully connected layers. The convolution operation shares the weight across the input that enables a CNN to learn features with a small number of parameters. Pooling is a downsampling method that helps to reduce the dimensions of the feature maps. Pooling helps to manage overfitting, reduce the computational cost by decreasing the number of weights in the model. CNNs are successfully applied to solving atmospheric dynamical phenomenons. In \cite{mudi2017seg, khjykkp2019dtfe} authors have used CNN to predict the TCs trajectories. In \cite{de_B_zenac_2019}, physical processes knowledge is combined with CNN to predict the SST. In \cite{swjjrlatpdm2020asotisri, crxwxac2019ahcltff} CNNs are used to predict the formation and intensity of a TC.

\subsection{Model Description}

\begin{figure}[!h]
    \centering
    \includegraphics[width = 0.5\columnwidth, height = 15cm]{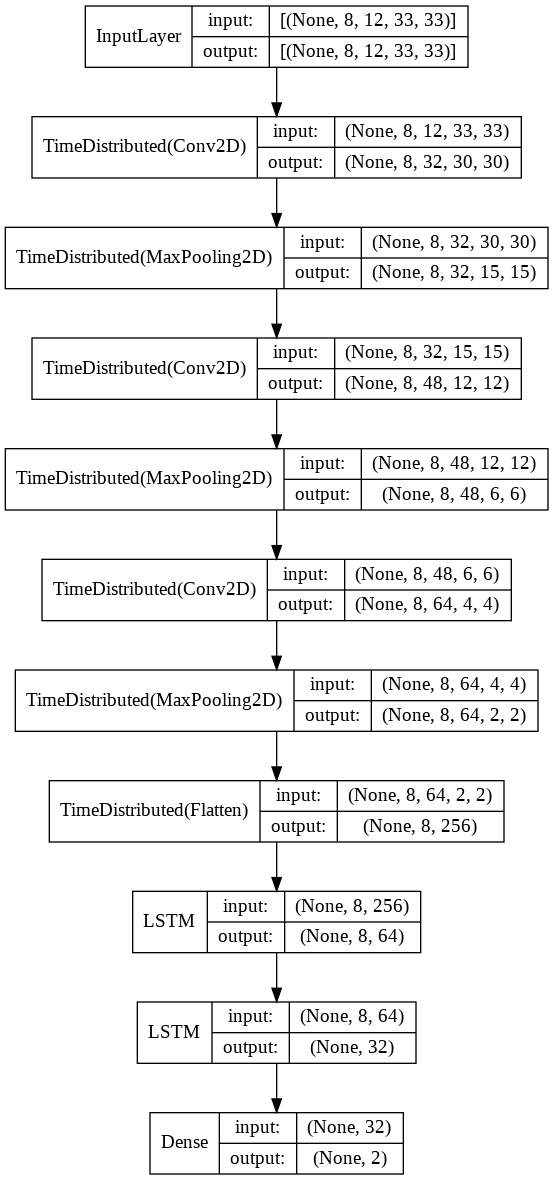}
    \caption{Model representation for latitude/longitude prediction for $T$ = 8}
    \label{fig2}
\end{figure}

We have used a combination of CNN and LSTM to capture the spatial and temporal aspects of our prediction problem.  We come up with a model that works well for each ocean basin. The structure of the model for location prediction in terms of latitude and longitude is described in the Figure~\ref{fig2} for $T = 8$.  The model for time prediction is the same except that the last dense layer's output size is one. The model is implemented in Keras API \cite{chollet2015keras} which uses underlying low-level language TensorFlow \cite{tensorflow2015-whitepaper}. For $T = 8$, one training point consists of eight sequential inputs of shape $12\times 33 \times 33$, where 12 represents the number of channels (nine channels of wind and geopotential fields, one of SST, two of latitudes and longitudes). We extracted the features corresponding to each time point of the input using the $\operatorname{TimeDistributed}$ layer of Keras, which are further fed into LSTM layers as shown in Figure~\ref{fig2}. 

\subsection{Training}

For training purposes, the dataset is divided into three parts training, validation, and test set in the ratio of 60:20:20.  We tried different configurations and hyperparameters for our model and choose the one that gives the best result on validation data after taking the issue of overfitting into account. Finally, we have reported the 5-fold test accuracy of our model. We trained the model for $T$ equals to 4, 6, and 8; increasing $T$ further does not lead to improved accuracy.  The input features are scaled using Standard Scaler of Scikit learn library \cite{scikit-learn}, which is given by $f(x) = \frac{x-\mu}{\sigma}$, where $\mu$ is the mean and $\sigma$ is the standard deviation. The target variables are also scaled in case of landfall's latitude and longitude prediction but not for time prediction.  The model uses the mean square error (MSE) as the loss function. We have reported the model performance in terms of root mean square error (RMSE) and mean absolute error (MAE), which are defined as follows:
\[
   \operatorname{MSE} = \frac{1}{n}\sum_{i=1}^{n}(y_{i} - \bar{y_{i}})^{2}, \quad \operatorname{RMSE} = \sqrt{MSE}, \quad
\operatorname{MAE} = \frac{1}{n}\sum_{i=1}^{n}\left | y_{i} - \bar{y_{i}} \right |,
\] where $y_i$ is the actual value and $\bar{y_i}$ is the predicted value.
The model uses the Adam \cite{kdj2014adam} optimizer with a default learning rate $0.001$ to minimize the loss function. Convolution layers and LSTM layers uses the activation function $\operatorname{ReLU}(x) = \max(0, x)$ \cite{relu}.  The model uses a total of $100$ epochs. We run our experiments on Nvidia Tesla V100 GPU with 16 GB RAM. The model takes approximately 30 to 45 minutes (depending on the ocean basin) to complete $100$ epochs. 


\section{Results and Comparison}\label{results}

Our model, at any time $t$ during the progress of a TC, takes $T$ number (($T$-1)*3 h) of data and predicts the landfall's latitude, longitude, and time. For example, if $T = 8$ and a particular TC is at the $t = 42h$ during its progression, then using the data between the time $t = 42-21 = 21$ and $t$. the model predicts the landfall's characteristics. To avoid any bias when time $t$ is very close to landfall's time, the model predicts only for $t \leq L-12$, where $L$ is the landfall's time, which means we are remaining at least 12 hours away from landfall while predicting. For each ocean basin and for different values of $T = 4, 6, 8$ (9h, 15h, 21h). we have reported the size of the dataset and the 5-fold accuracy on the test dataset in terms of RMSE and  MAE along with standard deviation in Table~\ref{time}. From the predicted latitude and longitude, the distance error between actual landfall location and predicted location is also reported in Table~\ref{time}. 

\begin{table}[!h]
\centering
\caption{5-fold accuracy on Test data of landfall's location and time prediction for different  $T$}
\label{time}
\begin{tabular}{|p{1.3cm}|p{0.9                    cm}|p{1cm}|p{1cm}|p{1cm}|p{1cm}|p{1cm}|p{1cm}|p{1cm}|p{1.3cm}|} 
\hline
\multirow{2}{4em}{\textbf{Ocean Basin}} & \multirow{2}{4em}{\textbf{T (hours)}} & \multirow{2}{4em}{\textbf{Dataset Size}} & \multicolumn{3}{|c|}{\textbf{RMSE (std)}} & \multicolumn{4}{|c|}{\textbf{MAE (std)}} \\
\cline{4-10}
 & &  & \textbf{Lati} degree  & \textbf{Long}  degree & \textbf{Time} hours & \textbf{Lati} degree & \textbf{Long}  degree & \textbf{Time} hours & \textbf{Distance} KM \\
\hline 
\multirow{3}{4em}{\textbf{North  Indian}} & 4 (9) & 5060 & 0.58 (0.05) & 1.03 (0.09) & 10.46 (1.40)& 0.38 (0.02)& 0.70 (0.05)& 7.33 (1.50)& 93.78 (6.71) \\
\cline{2-10}
 & 6 (15) & 4660 & 0.53 (0.13) & 0.90 (0.13) & 8.98 (3.47)& 0.36 (0.12)& 0.59 (0.10)& 6.05 (2.38)& 81.92 (18.62) \\
\cline{2-10} 
 & 8 (21) & 4284 & 0.40 (0.04)& 0.76 (0.06)& 6.72 (0.58) & 0.26 (0.02)& 0.50 (0.03) & 4.71 (0.54) & 66.18 (2.87)\\
\hhline{==========}

\multirow{3}{4em}{\textbf{South  Indian}} & 4 (9) & 9441  & 0.53 (0.03) & 1.78 (0.22)& 13.17 (0.59) & 0.36 (0.02)& 1.31 (0.15) & 8.32 (0.31) & 150.78 (15.55) \\
\cline{2-10}
 & 6 (15) & 8886 & 0.46 (0.03)& 1.44 (0.06)& 11.34 (0.84) & 0.30 (0.03) & 1.06 (0.04)& 7.24 (0.42)& 123.32 (5.24) \\
\cline{2-10} 
 & 8 (21) & 8353 & 0.42 (0.02)& 1.42 (0.18)&  9.63 (0.96) & 0.27 (0.03) & 1.05 (0.10) & 6.04 (0.45) & 119.96 (12.22) \\
\hhline{==========}
\multirow{3}{4em}{\textbf{East Pacific}} & 4 (9) & 3505 & 0.70 (0.23) & 1.49 (0.33) & 10.40 (1.03) & 0.52 (0.18) & 1.08 (0.24) & 7.17 (0.80)& 133.43 (33.42) \\
\cline{2-10}
 & 6 (15) & 3276 & 0.62 (0.04)& 1.33 (0.24)& 9.89 (2.17)& 0.46 (0.03) & 0.95 (0.10)& 6.99 (1.74)& 117.8 (9.76)\\
\cline{2-10} 
 & 8 (21) & 3056 & 0.52 (0.03) & 1.26 (0.16) & 11.28 (3.69) & 0.37 (0.02)& 0.93 (0.03)& 8.20 (2.96)& 110.48 (2.86) \\
\hhline{==========}

\multirow{3}{4em}{\textbf{South  Pacific}} & 4 (9) & 4885 & 0.89 (0.09) & 2.12 (0.14) & 17.44 (2.70) & 0.55 (0.05)& 1.50 (0.14)& 11.30 (2.19)& 179.03 (17.26) \\
\cline{2-10}
 & 6 (15) & 4520 & 0.89 (0.19) & 2.30 (0.57) & 13.24 (1.60)& 0.57 (0.14)& 1.57 (0.33)& 7.69 (0.65)& 188.81 (37.51)\\
\cline{2-10} 
 & 8 (21) & 4182 & 0.72 (0.10)& 1.67 (0.23)&  10.07 (1.77)& 0.44 (0.07)& 1.23 (0.20) & 6.74 (1.04) & 144.80 (22.44)\\
\hhline{==========}

\multirow{3}{4em}{\textbf{West  Pacific}} & 4 (9) & 34874 & 1.34 (0.34) & 1.72 (0.37) & 10.84 (0.30)& 0.88 (0.24) & 1.15 (0.26) & 7.44 (0.41)& 164.17 (40.08)\\
\cline{2-10}
 & 6 (15) & 32777 & 1.04 (0.16)& 1.40 (0.22)& 10.10 (2.14)& 0.72 (0.13) & 0.97 (0.16) & 7.22 (1.62) & 137.86 (24.28) \\
\cline{2-10} 
 & 8 (21) & 30791 & 0.79 (0.10)& 1.09 (0.10)& 8.12 (0.93)& 0.56 (0.08)& 0.76 (0.06) &  5.89 (0.72) & 108.0 (11.61)\\

\hhline{==========}

\multirow{3}{4em}{\textbf{North  Atlantic}} & 4 (9) & 9782 & 1.28 (0.12) & 2.42 (0.24) & 15.10 (4.20)& 0.84 (0.04) & 1.47 (0.11) & 10.47 (3.40)&  174.51 (10.68)\\
\cline{2-10}
 & 6 (15) & 8999 & 1.17 (0.15)& 2.13 (0.32) & 10.30 (1.26)& 0.79  (0.08) & 1.32 (0.14) & 6.58 (0.96) & 161.74 (15.36) \\
\cline{2-10} 
 & 8 (21) & 8276 & 1.05 (0.04)&  2.10 (0.34)& 10.69 (1.14)&  0.71 (0.03)& 1.38 (0.13) &  7.42 (0.98) & 158.92 (12.62)\\

\hline
\end{tabular}
\end{table}

From Table~\ref{time}, we can see that we can predict the landfall's location in ocean basins NI, SI, EP, SP, WP, and NA with a distance error of 66.18KM, 119.96KM, 110.48KM, 144.80KM, 108.0KM, and 158.92KM respectively for $T=8$ (21h) with a low standard deviation (std). If we look at the landfall's time prediction results, they are quite impressive. The model predicts the landfall's time in six ocean basins  NI, SI, EP, SP, WP, and NA with an MAE of 4.71h, 6.04h, 8.20h, 6.74h, 5.89h, and 7.42h respectively with low standard deviation. 

\begin{figure}[!h]
    \centering
    \includegraphics[width = 15cm, height = 5cm]{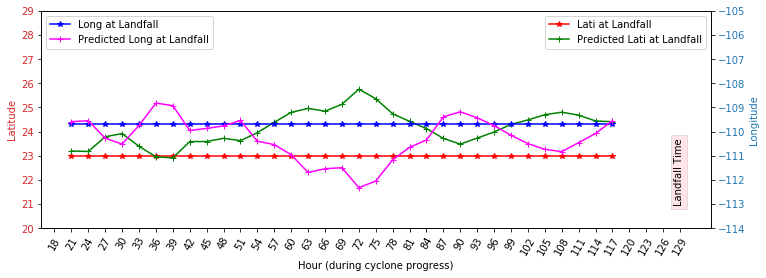}
    \caption{Latitude and Longitude prediction for hurricane BUD (2018) in East Pacific ocean for T = 8}
    \label{eploc}
\end{figure}

\begin{figure}[!h]
    \centering
    \includegraphics[width = 15cm, height = 5cm]{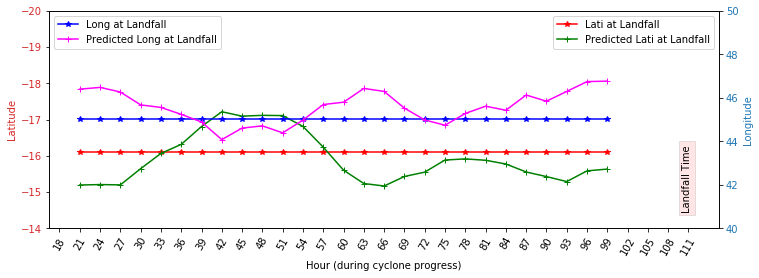}
    \caption{Latitude and Longitude prediction for cyclone BELNA (2019) in South Indian ocean for T = 8}
    \label{siloc}
\end{figure}

\begin{figure}[!h]
    \centering
    \includegraphics[width = 14cm, height = 5cm]{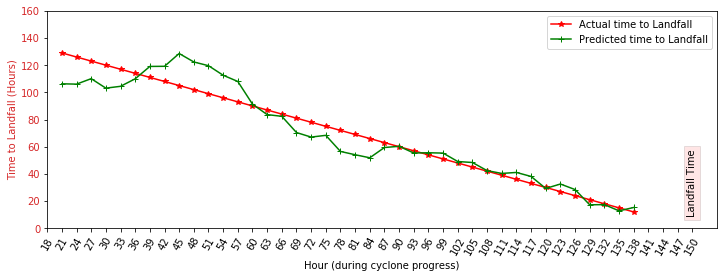}
    \caption{Time to Landfall prediction for Fani (2019) cyclone in North Indian ocean for T = 8}
    \label{nitime}
\end{figure}

\begin{figure}[!h]
    \centering
    \includegraphics[width = 14cm, height = 5cm]{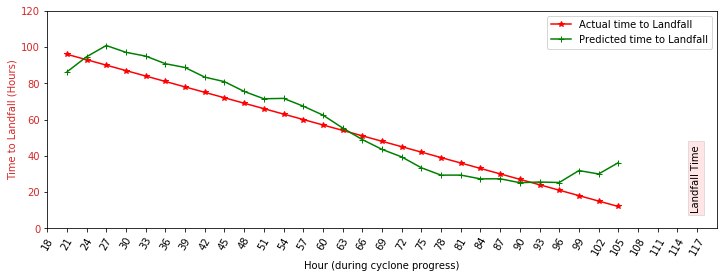}
    \caption{Time to Landfall prediction for NAKRI (2019) cyclone in West Pacific ocean for T = 8}
    \label{wptime}
\end{figure}

To further demonstrate the working of our model, in Figures~\ref{eploc} and \ref{siloc}, we have shown landfall's latitude-longitude prediction results for two cyclones BUD (2018) and BELNA (2019) in ocean basins EP and SI, respectively. In Figures~\ref{nitime} and \ref{wptime}, the landfall's time prediction results for TCs FANI (2019) and NAKRI (2019)  in ocean basins NI and WP, respectively, are shown. One can note that the model start predicting at time $t = 21$ and uses data of 21 hours between $t-21$ and $t$ to  predict at time $t$. All these named cyclones are not part of training data. 

\subsection{Comparison}

As per our knowledge, there does not exist any earlier work that predicts the landfall's location and time using reanalysis data. In the absence of directly related work, we will compare our results with closely related works \cite{gsygk2020tctufl, boussioux2020hurricane}. In Table~\ref{comp1}, the compassion between Fusion model \cite{gsygk2020tctufl} and our proposed model is shown. In \cite{gsygk2020tctufl} authors used TC data for 12 hours at a temporal resolution of 6 hours for track prediction with a lead time of 24 hours. A comparison with our model with $T=8$ is provided in Table~\ref{comp1}. One can see that we have achieved better results for a much harder problem with arbitrary lead time depending on the time between TC's initiation and its landfall. The authors of \cite{gsygk2020tctufl} have also reported the error range (std) 71KM for a subset of cyclones in the Atlantic ocean, while our model achieves an error range (std) 2.87KM - 22.44KM across all ocean basins. Also, the authors of \cite{gsygk2020tctufl} reported average performance on the test set for three runs of the model, while we have reported 5-fold performance of our model. All this makes our model more robust and reliable to use for practical purposes. 

\begin{table}[!h]
\centering
\caption{Comparison in terms of distance MAE (KM)}
\label{comp1}
\begin{tabular}{|p{2cm}|p{0.6                  cm}|p{0.6cm}|p{0.6cm}|p{0.6cm}|p{0.6cm}|p{0.6cm}|p{1.4cm}|p{1cm}|p{1.5cm}|p{1.5cm}|}
\hline
Model /Ocean Basin & NI (KM)& SI (KM) & EP (KM) & SP (KM) & WP (KM) & NA (KM) &No. of parameters ($*10^6$) & Training Time & Lead Time & Spacial Extent (Resolution) \\
\hline
Fusion (Track) \cite{gsygk2020tctufl} & 138.9 & 136.1 & 106.9 & 161.7 & 136.1 & 130.2 &  $\geq 2.27$ & 8h  & 24h & $25^\circ \times 25^\circ$ ($1^\circ \times 1^\circ$) \\
\hline
 Proposed Model (Landfall) & 66.1 & 119.9 & 110.4 & 144.8 & 108.0 & 158.9 & $0.157$ & $\leq$ 0.75h & TC dependent ($>=12h$) & $ 4^\circ\times 4^\circ$ ($0.25^\circ\times 0.25^\circ$)\\
\hline
\end{tabular}
\end{table}

\begin{table}[!h]
\centering
\caption{Track prediction MAE (std) in KM for HURR model and operational CLP5 model for 24 hour lead time}
\label{comp2hurr}
\begin{tabular}{|p{1cm}|p{3cm}|p{2cm}|p{2cm}|p{2cm}|}
\hline
 &{Model/Year} & {2017  (10 TCs) } & {2018  (15 TCs)} & {2019  (12 TCs)}  \\
\hline
\multirow{2}{2em}{\textbf{EP}} &{HURR-(viz, cnn/gru)} & 74 (40) & 69 (42) &  73 (45)\\
\cline{2-5}
 & {CLP5} & 114 (59) & 109 (61) &  133 (74) \\
\hline
\multirow{2}{2em}{\textbf{NA}} &{HURR-(viz, cnn/gru)} & 94 (61) & 113 (77) &  123 (89)\\
\cline{2-5}
 & {CLP5} & 189 (135) & 199 (118) &  207 (171) \\
\hline
\end{tabular}
\end{table}

\begin{table}[!h]
\caption{4/5 year (2015-2019) MAE reported by IMD for cyclones in NI ocean basin}
\label{imdData}
\centering
\begin{tabular}{ |p{3cm}|p{0.7cm}|p{0.7cm}|p{0.7cm}|p{0.7cm}| } 
  \hline
  \textbf{Lead Time (hours)} & 36 & 48 & 60 & 72 \\
 \hline
  \textbf{Landfall Time } &  4.96 &  5.53 & 6.8 &  9.6 \\ 
   \hline
 \textbf{Landfall Distance } &  42.84 &  78.08 & 92.6 & 112.5  \\ 
   \hline
\end{tabular}

\end{table}

In \cite{boussioux2020hurricane} authors use a combination of historical TC data, reanalysis data, and output from operational models and propose eight models for track prediction of TCs in NA and EP ocean basins for years 2017 to 2019. The model which close to our proposed model is $\operatorname{HURR-(viz, CNN/GRU)}$ which uses CNN-encoder and GRU-decoder \cite{gru}. The authors consider only those hurricanes in their study for which MSWS reaches 34 knots at some time $t_0$ and contains at least 60h of data after $t_0$. Our model does not have any such restrictions. In Table~\ref{comp2hurr}, we have shown the results for $\operatorname{HURR-(viz, CNN/GRU)}$ and the operational CLP5 model for EP and NA ocean basin. One can see that our results are not as good as that of $\operatorname{HURR-(viz, CNN/GRU)}$ but quite comparable with the operational model CLP5. One can notice that the standard deviation of our model is quite low in comparison to these two models. Here, we would again point out that we are not making a direct comparison here as our target prediction problem is different from the above-mentioned works but, at the same time, much more challenging and important.

We do not find any meteorological department across the world which reported the landfall prediction accuracy except the Indian Meteorological Department (IMD) on its website \cite{RMSC}. IMD has reported landfall's location error and time for a certain number of lead hours.  In Table~\ref{imdData}, we have reported the last 4/5 years (as per data availability) MAE achieved by IMD for landfall's location and time prediction. From Table~\ref{datasetDes}, we can see that in the NI ocean basin, the landfall occurs on average at 95.61 hours. Therefore, it is reasonable to compare our results with that of IMD for 72 lead hours. Clearly, here also, our model performs better than that of models used by IMD for both landfall's location and time. We have not included the results reported by IMD for earlier years, as errors are much higher. 

\section{Conclusion}\label{con}

We propose a model that can predict the landfall's location and time of a TC with high accuracy by observing a TC for 9h, 15h, or 21h at any time of its progression in the world's six ocean basins. The model took only 30 to 45 minutes for training and can predict the landfall characteristics within few seconds, which makes our model suitable for practical usage where the disaster managers can know the landfall location and time well in advance and can take preventive life saving and property saving measures well in time. Our model supports the case that deep learning models like CNN and LSTM can be utilized to predict challenging and complex prediction problems like the landfall of a TC.  One can further work in the direction of utilizing CNN with Attention and Transformers models for further improvement. One can also develop Consensus models to solve the proposed prediction problem.  As we are able to solve a complex landfall prediction problem using data over a small spatial extent with high resolution, it will be interesting to see to use the same kind of data for track or intensity prediction problems.

\bibliographystyle{unsrt}  

\bibliography{references}

\end{document}